\documentclass[review]{elsarticle}
\usepackage{hyperref}
\usepackage{amssymb}
\usepackage{amsmath}
\usepackage[font=footnotesize]{subfig}

\journal{Pattern Recognition}

\makeatletter
\define@key{Gin}{resolution}{\pdfimageresolution=#1\relax}
\makeatother

\begin{document}

\begin{frontmatter}

\title{Feature-based groupwise registration of historical aerial images to present-day ortho-photo maps}

\author{Sebastian Zambanini\fnref{myfootnote}}
\address{Computer Vision Lab\\
 Institute of Visual Computing and Human-Centered Technology, TU Wien\\
Favoritenstrasse 9/1931, A-1040 Vienna, Austria\\
sebastian.zambanini@tuwien.ac.at}
\fntext[myfootnote]{Corresponding author}

%
%

\begin{abstract}
In this paper, we address the registration of historical WWII images to present-day ortho-photo maps for the purpose of geolocalization. Due to the challenging nature of this problem, we propose to register the images jointly as a group rather than in a step-by-step manner. To this end, we exploit Hough Voting spaces as pairwise registration estimators and show how they can be integrated into a probabilistic groupwise registration framework that can be efficiently optimized. The feature-based nature of our registration framework allows to register images with a-priori unknown translational and rotational relations, and is also able to handle scale changes of up to 30\% in our test data due to a final geometrically guided matching step. The superiority of the proposed method over existing pairwise and groupwise registration methods is demonstrated on eight highly challenging sets of historical images with corresponding ortho-photo maps.
\end{abstract}


\end{frontmatter}


\section{Introduction}

Geolocalization of remote sensing images is an essential prerequisite for any further image analysis task as it provides a geometric mapping of the local image coordinates to the earth surface location. While modern aerial imaging systems are equipped with GPS and IMU to facilitate this task \cite{Mueller12}, historical aerial images usually lack this information and are thus less accurately pre-aligned. Specifically, orientation might be completely unknown and the image center position might be off to several kilometers, in contrast to modern aerial images whose geolocation error is typically less than 10 meters \cite{Ozcanli16}. Accurate geolocalization can be accomplished by setting image points and known ground control points in correspondence \cite{Wong07}, in which case the geolocalization task becomes a classical image registration problem \cite{Zitova03}, where the input image is aligned to a georeferenced target image. In such a setting, target images need to be easily accessible, and consequently either Ortho-Photo Maps (OPMs) \cite{Lin07} or road network maps \cite{Costea16,Mattyus16} are utilized. However, in any case, identifying correspondences is much more challenging when working with historical images, as due to the rougher pre-alignment the translational and rotational search space for corresponding features is larger and corresponding image points are exposed to temporal changes.

\begin{figure*}[ht!]
	\centering
(a)\includegraphics[height=6.9cm, resolution=10]{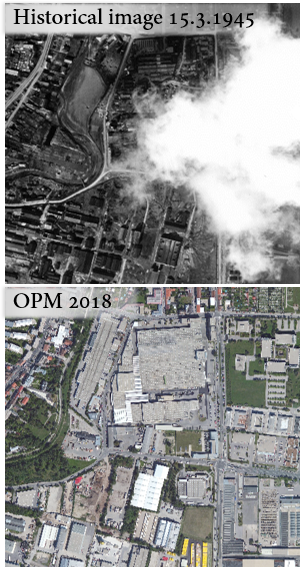}
(b)\includegraphics[height=6.9cm, resolution=10]{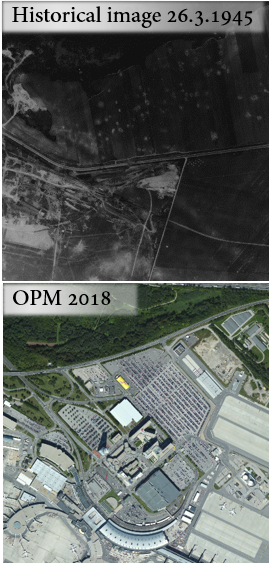}
(c)\includegraphics[height=6.9cm, resolution=10]{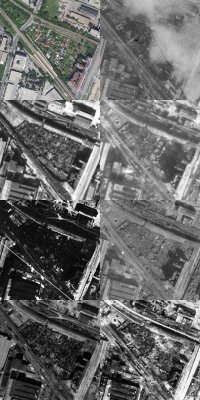}
		\caption{Challenges of the historical image material: pairs of already aligned historical-OPM image patches of a) Vienna’s 23rd district and b) the airport in Schwechat, Austria (right); c) Overall variation of geographic region appearance in seven time-separated historical images.}
		\label{fig:challenges}
\end{figure*}

In this paper, we tackle the problem of automatically registering historical aerial photographs to present-day OPMs. In particular, we address the case when multiple images from different times are available and propose to leverage the geometrical relationships of all images to jointly registering them to the reference image. Georeferencing single multi-temporal images of a particular region-of-interest is a typical use case when analyzing historical WW2 image material for UneXploded Ordnance (UXO) risks, e.g. by detecting bomb craters in the images \cite{Brenner18,Kruse18}. As shown in Figure \ref{fig:challenges}, these images pose specific challenges to the registration problem. In general, the images are gray scale only and might be affected by over- or underexposure, blurring or sensor noise. However, most notably, the registration can be hindered by the lack of corresponding structures between historical and present-day images. In urban areas, structures like streets and buildings may have been preserved, but image matching can be hindered by other factors such as cloud coverage (Figure \ref{fig:challenges}(a)). For other areas, the extensive construction of new buildings render the matching process impossible, as no corresponding image features exist anymore (Figure \ref{fig:challenges}(b)). Also among the historical images a high degree of variation is given, owing to various effects like destroyed structures, weather and illumination conditions or smoke (Figure \ref{fig:challenges}(c)).  Additionally, as for the historical images only the approximate image scale and geolocation of the image center is available, leveraging this prior information of unknown quality can only be done in a tentative manner.

Traditionally, image registration methods follow either a \emph{local} or \emph{global} approach \cite{Wen08}. Global methods aim to use a global similarity metric like mutual information \cite{Zhilkin08} to find optimal transformation parameters. Although being effective in the sense that the whole image information is exploited to guide the registration, these techniques suffer from a high computational load and local minima trapping and are thus more suitable for the fine registration of already roughly aligned image pairs \cite{Liang14}. In contrast, with local techniques the estimation of transformation parameters is based on identified matching keypoints between the images. By using local image detectors and descriptors that are insensitive to geometric transformations \cite{Lowe04,Bay08,Heikkila09}, a pre-alignment is not needed. Additionally, by design, these techniques are less affected by structurally different image parts, e.g. resulting from temporal changes. Consequently, the key challenge of a feature-based image registration task like the one addressed in this paper is to robustly establish local correspondences between the images. A major aspect is to use both discriminative and insensitive local image descriptors, but still using the appearance information alone is too ambiguous to identify correspondences in a robust manner. Therefore, geometric constraints are typically used to select or verify the correspondences \cite{Wu15}. A common methodology is geometric verification, whose idea is to check the geometric consistency of the initially determined correspondences. A possible verification constraint is that the correspondences follow a common global geometric transformation. The RANdom SAmple Consensus (RANSAC) \cite{Fischler81}
scheme repeatedly takes random subselections of the feature correspondences and checks how many correspondences support the global transformation estimated from the chosen samples. The general concept of treating spatial verification as a postprocessing step that filters out the erroneous ones from the overall set of putative matches has been developed further by many researchers \cite{Ma15,Raguram08,Song14,Cheng09}. However, these methods demands the initial matching process to be discriminative enough to produce a statistically significant inlier ratio. An alternative methodology has been proposed as to iteratively refine and enrich the matches based on the initially identified strongest ones \cite{Yang07,Han14,Hu16,Ma18}. Still, the basic assumption is that the discriminative power of the local descriptors is strong enough to provide a valid initialization. Graph matching methods \cite{Cho12,Yang17} aim to optimize the matching process by jointly considering local appearance and geometric constraints, but also have to rely on a small set of initial candidate matches for reasons of computational tractability. 

In the past, a lot of research has been conducted to improve feature matching at the descriptor level, e.g. by extracting multiple types of descriptors that embody different strengths and thus allow a more versatile matching \cite{Hu15, Mishkin15}. Designing specialized descriptors for specific types of image changes has also been vastly addressed, such as different modalities \cite{Kelman07}, illumination changes~\cite{Zambanini13} or wide baselines~\cite{Yu11}. Line features have been reported to have stronger repeatability than point features in the area of remote sensing image registration~\cite{Nagarajan16,Zhao18}. More recently, feature learning has been followed rather than hand-crafting them~\cite{Kumar16,Simo-Serra15,Ye18}. 

However, although guided matching strategies and well-adapted descriptors methods can be partly effective, they do not solve a fundamental problem common to historical-to-OPM registration, i.e. that possibly large parts of the images are not matchable due to temporal changes. This issue, together with the other challenges discussed above, lead to inlier ratios of less than 5\% when using a standard feature matching approach (see Section \ref{sec:standard}).

Therefore, we propose to handle the registration process in a \emph{groupwise} manner. Instead of registering images stepwise to each other or to the reference OPM, we cast the registration as an optimization problem, where the optimal global configuration is found by considering the likelihoods of all historical-to-OPM and historical-to-historical spatial transformations. Registering groups iof images jointly instead of in a pairwise fashion has been previously identified to provide higher registration accuracy and robustness in various computer vision domains, e.g. medical imaging \cite{Huizinga16} or remote sensing \cite{Arandjelovic15}. Typically, these methods follow the global registration paradigm and aim to optimize a joint global image similarity metric, e.g. mutual information. Consequently, they assume already pre-aligned images \cite{Orchard10,Spiclin12,Wachinger13} and can thus not be applied to our type of problem. Arandjelovic et al. \cite{Arandjelovic15} explicitly address the groupwise registration of aerial images from different time stamps, but their method is restricted by describing geometric change between image pairs by translation only. 

Other works in multiple aerial image registration usually assume that the images stem from the same source (e.g. a flight sequence) and that pairwise registrations can be constrained by the sequence order \cite{Kekec14,Xia15}. For unordered image sets, registration is typically achieved by image stitching methods \cite{Xiang18}, where optimal reference images and linkages are identified by evaluating all pairwise registration attempts \cite{Xia17,Elibol13}. Once an initial solution has been identified, bundle adjustment techniques \cite{Marzotto04,Konolige10} can be used to jointly reduce the registration error. With such methods, the strengths of feature-based registration can be exploited, i.e. a higher robustness in cases of largely unmatchable image parts and more extensive geometric differences between image pairs. However, their premise is that pairwise registrations are reliable enough in the first place to find all-encompassing linkages and finally refine overall registration consistency. 

In contrast to previous methods, our method is not a pure refinement of already roughly aligned images and is by concept able to handle a-priori unknown translation and rotation between images. It builds upon a recently proposed Hough voting mechanism \cite{Zambanini17} that transfers local feature correspondences to probabilities of spatial transformations between images. Therefore, our method is a feature-based groupwise registration that does not aim to find optimal registration paths in previously determined pairwise registrations, but rather exploits the Hough voting space as pairwise registration estimator to find the most likely global registration solution.

The remainder of this paper is organized as follows. In Section \ref{sec:method}, the proposed methodology is described in detail, including the Hough voting space as pairwise registration estimator, the sequential global optimization method and the final locally guided matching. Experiments on a comprehensive dataset of historical aerial images are reported and discussed in Section \ref{sec:results}. In Section \ref{sec:conclusions}, concluding remarks are given.

\section{Methodology}
\label{sec:method}

\begin{figure}[ht!]%
	\centering
\includegraphics[width=\textwidth, resolution=100]{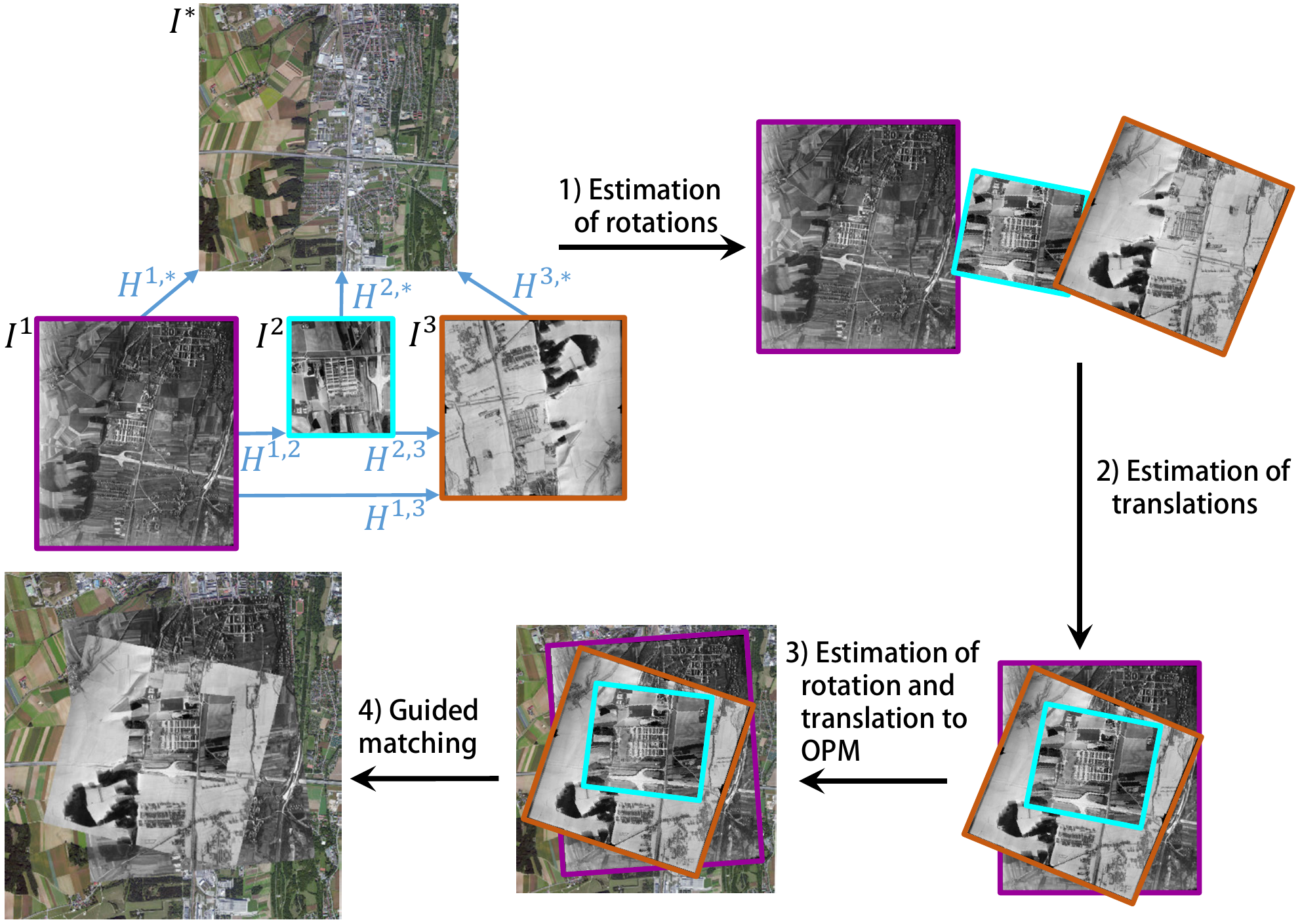}%
\caption{Overview of our probabilistic method for groupwise registration of aerial historical images to a reference OPM $I^*$ (in this example, three historical images $I^1$, $I^2$ and $I^3$ are used). We start with the computation of pairwise Hough voting spaces $H^{k,l}$ that encode the probability of all rigid transformations between the images. The joint max-likelihood optimization is carried out sequentially, by optimizing for 1) rotations among historical images, 2) translations among historical images and 3) translation and rotation to OPM. These rigid relations are finally used to 4) geometrically guide feature matching.}
\label{fig:overview}%
\end{figure}

We propose a probabilistic method for groupwise image registration where probabilities of pairwise registrations are evaluated and summed up to find the optimal global groupwise registration. In this case, pairwise probabilities need to be evaluated continuously and calculating them online is not feasible. Therefore, we exploit the idea of a Hough voting space as pairwise registration estimator~\cite{Zambanini17}, as described in Section \ref{sec:hough}. After Hough voting spaces are calculated for each image pair, they store the probabilities for all transformations between the image pair, which can be efficiently looked-up during the optimization procedure described in Section \ref{sec:groupwise}. To keep the optimization manageable, pairwise image relations are simplified to translation and rotation. In order to account for a more realistic model of pairwise image relations, we retrieve the homography relations between historical images and the OPM in a final guided matching step, as described in Section \ref{sec:guidedmatching}. An overview of our method and its individual processing steps is shown in Figure \ref{fig:overview}.

\subsection{Hough voting space as pairwise registration estimator}
\label{sec:hough}

Fundamental to probabilistic groupwise registration is an estimator that returns the probability of a given transformation between two images. While with global registration the similarity score of the metric used can be directly used for this purpose, there is no straightforward method with feature-based registration. In this paper, we thus use a Hough voting space \cite{Zambanini17} as pairwise registration estimator, where the similarity scores of matched features with similar geometric relation are accumulated. Once this 3D Hough space  $H$ between two images is calculated, it stores the probabilities of all possible transformations between these images, i.e. $H(\mathbf{v},\gamma)$ gives the probability of a rigid transformation with translation vector $\mathbf{v}=(vx,vy)$ and rotation angle $\gamma$.  We neglect scale in this model as the images are considered to be scale-normalized, since the scales are typically given in remote sensing images and can also be estimated in the case of historical WWII aerial images by means of the known aircraft altitudes and the focal lengths of the cameras used \cite{Campbell11}. However, given aircraft altitudes are only approximated in this case, as during combat stable altitudes could not be guaranteed. Therefore, the proposed method  accommodates for scale changes in the final guided matching process. This way, typically small scale normalization errors of up to 30 \% can be properly handled, as shown in the experiments.

In order to build $H$ between two images, the local geometry of matched features is transferred to weighted votes. Local image features such as SIFT \cite{Lowe04} deliver descriptor vectors $\mathbf{d}_i$ together with local feature frames $\mathbf{f}_i= (x_i,y_i,\sigma_i,\theta_i)$. Local geometry is described here by the feature location $(x_i,y_i)$ relative to the image center, the scale $\sigma_i$ and the orientation $\theta_i$. To compute the Hough space $H^{k,l}$ between the images $I^k$ and image $I^l$ of a group, we first compute similarities\footnote{Generally, the defined method of converting distances to similarities could be problematic in terms of numerical stability when distances are (close to) $0$. However, local feature distances used in this work are typically in a range of $\left[50,750\right]$} between all features $\mathbf{d}^k_i$ and $\mathbf{d}^l_j$ as $s^{k,l}_{i,j}=(\lVert \mathbf{d}^k_i - \mathbf{d}^{l}_j \rVert_2)^{-1}$. Then, a subset $\mathcal{M}^{k,l}$ containing the matches with highest similarity is taken, $\mathbf{m}^{k,l}_{i,j} = (\mathbf{f}^k_i,\mathbf{f}^{l}_j) \in \mathcal{M}^{k,l}$. Each $\mathbf{m}^{k,l}_{i,j} \in \mathcal{M}^{k,l}$ votes for a rigid transformation in the 3D Hough space $H^{k,l}(vx({\mathbf{m}^{k,l}_{i,j}}), vy(\mathbf{m}^{k,l}_{i,j}), \gamma(\mathbf{m}^{k,l}_{i,j}))$, with

\begin{equation}
	\gamma(\mathbf{m}^{k,l}_{i,j}) = \theta^k_i - \theta^l_j,
\end{equation}

\begin{equation}
	\begin{pmatrix} vx({\mathbf{m}^{k,l}_{i,j}})\\ vy({\mathbf{m}^{k,l}_{i,j}}) \end{pmatrix} = \begin{pmatrix}x^{l}_j\\ y^{l}_j \end{pmatrix} - \begin{pmatrix} \cos{\gamma(\mathbf{m}^{k,l}_{i,j})} & -\sin{\gamma(\mathbf{m}^{k,l}_{i,j})}\\ \sin{\gamma(\mathbf{m}^{k,l}_{i,j})} & \cos{\gamma(\mathbf{m}^{k,l}_{i,j})}  \end{pmatrix} \cdot \begin{pmatrix}x^{k}_i\\ y^{k}_i \end{pmatrix}.
\end{equation}

$H^{k,l}$ is initialized with zeros and updated as 

\begin{equation}
	H^{k,l}(vx({\mathbf{m}^{k,l}_{i,j}}), vy(\mathbf{m}^{k,l}_{i,j}), \gamma(\mathbf{m}^{k,l}_{i,j})) = H^{k,l}(vx({\mathbf{m}^{k,l}_{i,j}}), vy(\mathbf{m}^{k,l}_{i,j}), \gamma(\mathbf{m}^{k,l}_{i,j})) + s^{k,l}_{i,j}
	\label{eq:hough_update}
\end{equation}

\noindent for all $\mathbf{m}^{k,l}_{i,j} = (\mathbf{f}^k_i,\mathbf{f}^{l}_j) \in \mathcal{M}^{k,l}$. After all votes have been cast using Eqn. \ref{eq:hough_update}, $H^{k,l}$ is normalized such that $\sum_{\forall vx,vy,\gamma}{H^{k,l}(vx,vy,\gamma)}=1$.

\begin{figure}[b!]
	\centering
\includegraphics[width=\columnwidth]{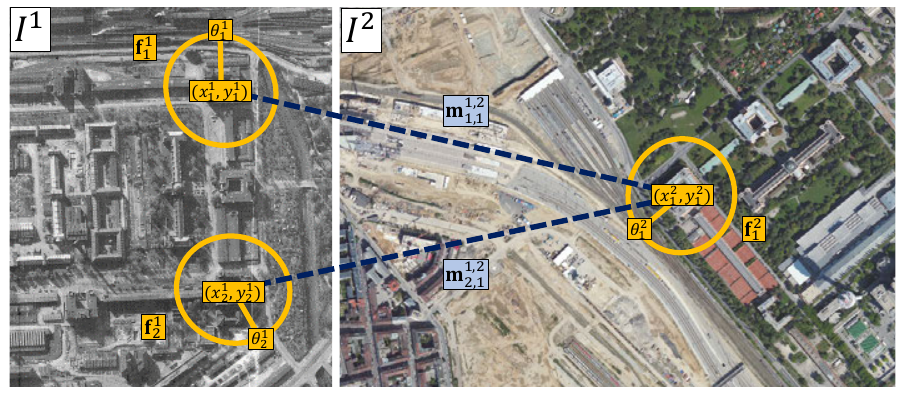}
		\caption{An illustration of two features $\mathbf{f}^1_1$ and $\mathbf{f}^1_2$ in image $I^1$ and a feature $\mathbf{f}^2_1$ in image $I^2$ and how their local geometry contribute to the Hough space $H^{1,2}$ by the matches $\mathbf{m}^{1,2}_{1,1}$ and $\mathbf{m}^{1,2}_{2,1}$. Here, features are shown as circles whose centers indicate the feature position $(x,y)$ and the line inside the circles indicate feature orientation $\theta$. Matches are represented by dashed blue lines.}
		\label{fig:illu_vote}
\end{figure}

An illustration of this process is shown in Figure \ref{fig:illu_vote}. Here we see two matches between two images $I^1$ and $I^2$, one being correct ($\mathbf{m}^{1,2}_{1,1}$), the other being incorrect ($\mathbf{m}^{1,2}_{2,1}$). The correct match votes with its similarity score $s^{1,2}_{1,1}$ for the correct translation and rotation (approx. $135^\circ$) between the images, while the incorrect match votes for an incorrect translation and rotation. Notably, given that the vast majority of matches can be assumed to be incorrect, a single match provides only a weak and unreliable evidence for the geometric relation of the two images. However, when using a high number of matches (100.000 in our case), the determined probability of the correct transformation can be expected to be at the maximum or at least above average. It is shown in Section \ref{sec:comparison1} that in fact for about 25\% of historical-to-OPM registrations the correct transformation is at the maximum, while the remaining estimated probabilities of pairwise historical-to-historical and historical-to-OPM relations are strong enough to determine the correct transformation for all image pairs with our proposed probabilistic groupwise registration procedure. 

As in \cite{Zambanini17}, we also apply a correspondence zoning procedure that allows only the strongest vote in $\mathcal{M}^{k,l}$ between two specific image areas to be added to the Hough space, in order to avoid a disproportional influence of votes caused by nearby matches. Naturally, the update of Eqn. \ref{eq:hough_update} needs to done in quantized Hough space. For the rotation $\gamma$, an interval of $2\pi/18$ is used and the similarities are bilinearly interpolated to distribute their values to adjacent bins. For the translation $\mathbf{v}$, a quantization interval of one pixel (corresponding to 1m) is used, and missing values are reconstructed by means of a Gaussian filter.

\subsection{Probabilistic groupwise registration}
\label{sec:groupwise}

In this section, we show how the pairwise registration estimations $H^{k,l}$ can be leveraged to solve the registration problem in a groupwise manner. The global fitness function that needs to be maximized is described in Section \ref{sec:fitness}. In Section \ref{sec:optimization}, we present a sequential method to efficiently solve this fitness function.

\subsubsection{Fitness function}
\label{sec:fitness}

The goal of groupwise registration is to balance pairwise registrations in order to find the most likely joint solution. We therefore cast the registration of historical aerial images as an optimization problem where $N$ historical images $I^1 \dots I^N$ need to be registered to a reference OPM $I^*$. For each $I^k$ we seek the optimal rigid transformations $\mathbf{T}_k = (vx_k,vy_k,\gamma_k)$ to $I^*$ given the precomputed Hough spaces $H^{k,l}(\mathbf{T})$ that deliver the likelihood of the transformation $\mathbf{T}$ between the images $I^k$ and $I^l$. With groupwise registration, the fitness function to be maximized considers the overall set $\mathcal{T} = \{\mathbf{T}_1,\dots,\mathbf{T}_N\}$ of transformations,

\begin{equation}
J(\mathcal{T}) = \underbrace{\sum_{k=1}^N{H^{k,*}(\mathbf{T}_k)}}_{\substack{Direct \ relations}} + \ \ \ \underbrace{\sum_{k=1}^N{\sum_{l=2}^N{w(k,l) \cdot H^{k,l}(c(\mathbf{T}_k,\mathbf{T}_{l^{-1}}))}}}_{\substack{Indirect \ relations}}.
\label{eq:J}
\end{equation}

\begin{figure}[b!]
	\centering
\includegraphics[width=0.6\columnwidth]{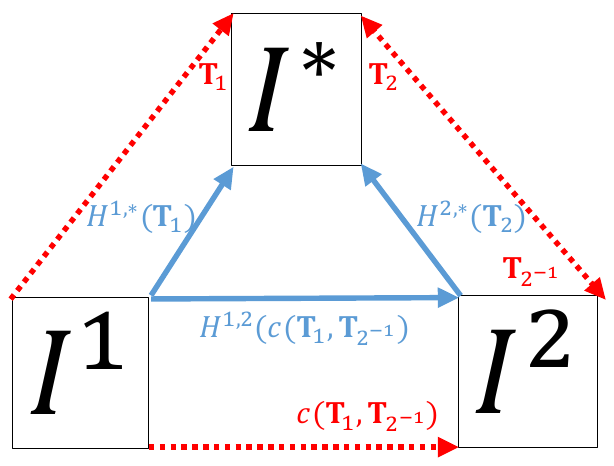}
		\caption{Illustration of direct and indirect relations considered when evaluating the likelihood of the transformations $\mathbf{T}_1$ and $\mathbf{T}_2$ of the images $I^1$ and $I^2$ to the reference OPM $I^*$. Transformations are depicted as red dashed arrows, and their corresponding likelihoods as solid blue lines.}
		\label{fig:J}
\end{figure}

This fitness function consists both of a \emph{direct} and  \emph{indirect} relations term. The direct term considers only the relations of the historical images to the OPM. Optimizing this term alone could be easily done by taking the maximum values of all $H^{k,*}$ and would correspond to $N$ simple independent historical-to-OPM registrations. Thus, the groupwise approach is attributed to the indirect term, whose role is to evaluate pairs of transformations by means of the likelihood of the corresponding historical-to-historical registrations, as illustrated in Figure \ref{fig:J}. Here, the composite transformations $c(\mathbf{T}_k,\mathbf{T}_{l^{-1}})$ of $\mathbf{T}_k$ and the inverse transformation $\mathbf{T}_{l^{-1}}$ are computed as

\begin{align}
c(\mathbf{T}_k,\mathbf{T}_{l^{-1}}) = \Big( &\nonumber (vx_k-vx_l)\cos{\gamma_l}  + (vy_l-vy_k)\sin{\gamma_l}, \\
                           &\nonumber (vy_k-vy_l)\cos{\gamma_l} + (vx_k-vx_l)\sin{\gamma_l}, \\ 
													&\gamma_k-\gamma_l) \Big)
\label{eq:c}
\end{align}

The function $w(k,l) \in \{0,1\}$ of Eqn. \ref{eq:J} is a binary indicator that determines if the relation between the images $I^k$ and $I^l$ should be included. In our case, all pairwise relations are considered, i.e. $w(k,l)=1 \ \forall \ k \neq l$.

\subsubsection{Sequential global optimization}
\label{sec:optimization}

Directly optimizing Eqn. \ref{eq:J} is problematic due to the high-dimensional non-convex fitting function: for $N$ images, we have to optimize $3N$ parameters and false local matches tend to produce a vast amount of local maxima in the Hough voting space. Therefore, we propose to solve the optimization problem sequentially, where we break down the overall task to a series of subtasks with a lower dimensional search space. In each subtask, a subset of parameters is optimized and fixed for the next subtask. The actual optimization in each subtask is carried out by a basic solver. In our case, we use Particle Swarm Optimization (PSO) \cite{Nouaouria13} as it provides a straightforward way to adapt the optimizer to the reliability of initial solutions by controlling the overall number and starting points of the particles.

In our approach, we account for the nature of groupwise historical-to-OPM by leveraging the fact that the pairwise historical-to-historical estimators $H^{k,l}$ are far more reliable than the  historical-to-OPM estimators $H^{k,*}$. We therefore first solve for the rotations among the set of historical images (N-1 parameters), followed by the determination of their pairwise translations (2(N-1) parameters). Finally, the translation and rotation to the OPM is determined (3 parameters).

\paragraph{Rotations among historical images}

Solving for rotation only demands for a rotation estimator between images. For this purpose, we can easily compress the 3D rigid transformation estimator $H^{k,l}$ to a 1D rotation estimator $H^{k,l}_R$ by max-pooling over the translation dimensions:

\begin{equation}
H^{k,l}_R(\gamma) = \max_{\forall vx,vy}{H^{k,l}(vx,vy,\gamma)},
\end{equation}

\noindent followed by a normalization such that  $\sum_{\forall \gamma}{H_R^{k,l}(\gamma)}=1$. Analog to Eqn. \ref{eq:J}, the fitness function to be maximized consists of direct and indirect relations,

\begin{equation}
J_R(\mathcal{R}) = \sum_{k=2}^N{H_R^{k,1}(R_k)} + \ \ \sum_{k=2}^N{\sum_{l=3}^N{w(k,l) \cdot H_R^{k,l}(c(R_k,R_{l^{-1}}))}}.
\end{equation}

Here, by definition, the first historical image is the reference image, i.e. the set of rotations $\mathcal{R} = \{R_2,\dots,R_N\}$ of all other images to this historical image needs to be found. 

Although this optimization subtask is easier to solve due to the lower number of parameters ($N-1$ instead of $3N$), the optimizer can still easily be stuck in local maxima. Therefore, proper initialization is essential for a successful optimization identifying the global maximum. We propose a simple but effective greedy algorithm for determining the initialization values  $\mathcal{R}^\prime = \{R_2^\prime,\dots,R_N^\prime\}$ based on the graph of rotation relations. In this graph, the images serve as nodes and rotation relations serve as edges. The edges are weighted by the inverse highest likelihood given in $H_R^{k,l}$, i.e. the relation between image $k$ and $l$ is weighted by $H_R^{k,l}(\widehat{R})^{-1}$, with $\widehat{R} = \arg \max{H_R^{k,l}}$. 

\begin{figure}[b!]%
	\centering
\includegraphics[width=0.8\columnwidth]{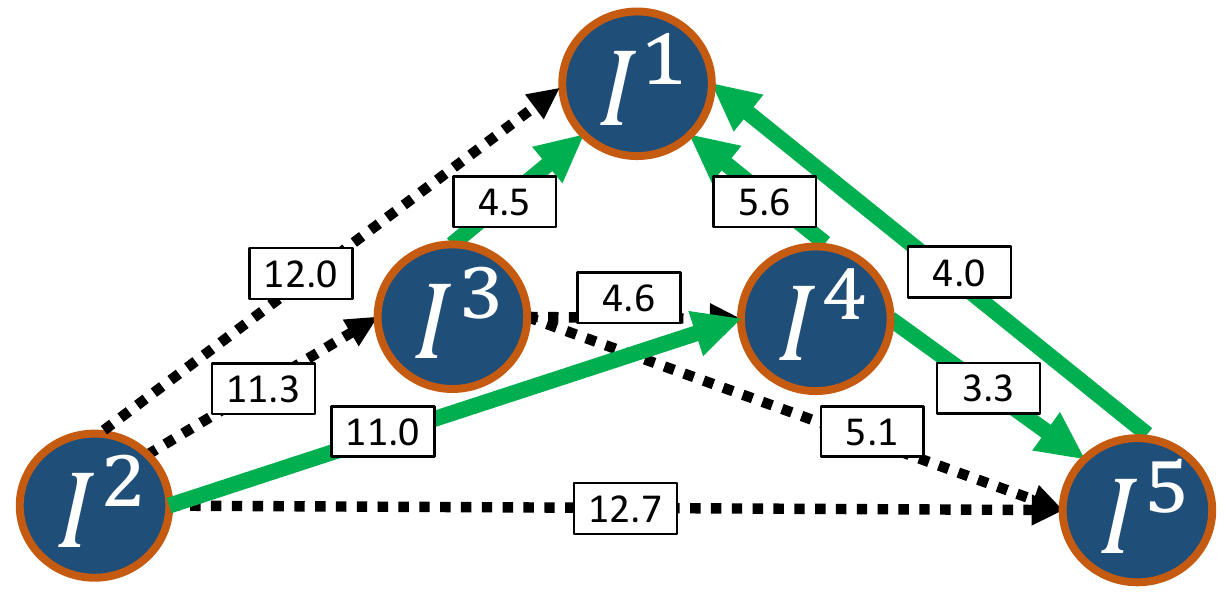}%
\caption{A graph of weighted rotation relations used to derive initialization values for optimizing rotations among historical images. The green solid edges represent the paths finally used.}%
\label{fig:graph}%
\end{figure}

An example for a graph constructed from five historical images is shown in Figure \ref{fig:graph}. The search for initialization values works as follows. We first sort the edges in ascending order. We then add these edges to a new graph and check after each addition if a shortest path from any image to $I^1$ is available (depicted by the green solid edges in Figure \ref{fig:graph}). If this is the case, the concatenated rotation to $I^1$ is stored as initialization value for this image, and the inverse average weights along the path as its confidence value. This way, we ensure that the initialization values are based on the most reliable rotation estimations among the group of historical images.

Once initialization values and corresponding confidence values have been estimated, this information needs to be exploited for the particle initialization of PSO. An inspection of the determined initialization values reveals that most of these values are already close to the correct solution, so we only change a specified percentage of initialization values with lowest confidence. Based on empirical tests, we use 150 particles for PSO, and for all (N-1)-dimensional particles the values of the dimensions belonging to the lowest  $30\%$ of confidences are randomly changed. For instance, in the example of Figure \ref{fig:graph} the rotation relations from $I^3$ to $I^1$ and $I^5$ to $I^1$ are the most confident ones, and hence both $R_3^\prime$ and $R_5^\prime$ are set to the maximum given in $H^{3,1}_R$ and $H^{5,1}_R$, respectively, and kept fixed for all PSO particles. $R_2^\prime$ and $R_4^\prime$ are randomly chosen for each of the 150 particles.

\paragraph{Translations among historical images}

Once the rotation relations among the historical images have been determined, we can simply derive 2D translation estimators $H_{\mathbf{V}}^{k,l}$ by applying the determined rotation values to $H^{k,l}$. Then, a set of translations $\mathcal{V} = \{\mathbf{V}_2,\dots,\mathbf{V}_N\}$ is sought that maximizes the direct and indirect translation relations among historical images,

\begin{equation}
J_{\mathbf{V}}(\mathcal{V}) = \sum_{k=2}^N{H_{\mathbf{V}}^{k,1}(\mathbf{V}_k)} + \ \ \sum_{k=2}^N{\sum_{l=3}^N{w(k,l) \cdot H_{\mathbf{V}}^{k,l}(c(\mathbf{V}_k,\mathbf{V}_{l^{-1}}))}}.
\end{equation}

Initialization is carried by the same greedy algorithm as for rotation estimation. For translation, initialization values are already close to the correct solution so we simply run PSO with 150 particles with the determined initializations, each particle being altered by a random value from a Gaussian distribution with a standard deviation of 3m.

\paragraph{Translation and rotation to OPM}

The previous two steps are responsible for estimating optimal rigid transformations among historical images. In the final step, their rigid transformation to the OPM is determined. In other words, we optimize only for the three parameters of $T_1$ while the other transformations $T_2,\dots,T_N$ used for the fitness function defined in Eqn. \ref{eq:J} are settled by the already fixed transformations $\{\mathcal{V},\mathcal{R}\}$ among the historical images. For initialization, we take for each historical image the most likely transformation to the OPM according to $H^{k,*}$ and initialize five particles with these values.

After this procedure, the found solution could be assumed be close to the global maximum. We thus are able to further refine the solution by jointly optimizing all $3N$ parameters by a quasi-Newton optimization \cite{Arora15} of  Eqn. \ref{eq:J}. 

\subsection{Final historical-to-OPM registration by geometrically guided matching}
\label{sec:guidedmatching}

Our proposed probabilistic framework for groupwise registration can be considered as a rough alignment of images, as the transformation between images is modeled by a rigid transform only. Therefore, we apply a final guided matching step that takes the estimated rigid transform to limit the search space for correspondences and leads to the estimation of a more general homography transform between images. However, as establishing correspondences directly between historical images and the OPM is error-prone due to possibly large changes in image content, we apply the same greedy graph search algorithm described in Section \ref{sec:optimization} to find reliable paths from an historical image to the OPM. The edges of the graph are thereby weighted by the inverse probability $H^{k,l}(\mathbf{\widetilde{T}}^{k,l})^{-1}$, with $\mathbf{\widetilde{T}}^{k,l}$ being the estimated transformation between images $I^k$ and $I^l$. Again, edges are sorted by their weights in ascending order and added to a new graph, each time checking for a possible path of a historical image to the OPM. If a path is found, relations between images connected by an edge are determined by the guided matching step, which works as follows:

\begin{enumerate}
	\item Detect keypoints using Difference-of-Gaussians \cite{Lowe04} in both images and align keypoint positions according to the given rigid transform between the images. Compute SIFT descriptors at the same canonical orientation according to the given rotation between the images.
	\item For each feature in the first image, find closest match in other image for all features with position distance and scale distance below a specified threshold. For the position, we use an empirically determined threshold of 500m. For the scale, only features with a scale ratio below $\varepsilon$ and above $\epsilon^{-1}$ are considered, with $\epsilon=1.4$.
	\item Apply RANSAC \cite{Fischler81} to the tentative matches to filter out outliers and estimate the homography transform between the images.
\end{enumerate}

\section{Experiments}
\label{sec:results}

In this section, we provide a thorough evaluation of our proposed methodology. Our dataset of historical aerial images from WWII is described in Section~\ref{sec:data}. In Section~\ref{sec:standard}, we first we perform some tests to demonstrate the limited use of the standard local feature matching approach \cite{Lowe04} for our data. In Section~\ref{sec:comparison1}, we compare our proposed groupwise registration method to common pairwise historical-to-OPM registration and additionally investigate the improvement achieved by guided matching. Section~\ref{sec:optimization_results} provides an empirical investigation of the proposed sequential global optimization procedure. A comparison to an existing state-of-the-art groupwise registration method is given in Section \ref{sec:comparison2}. 

\subsection{Test data and evaluation metric}
\label{sec:data}

We have compiled a dataset consisting of 8 reference OPM images from urban and non-urban areas in Austria\footnote{The dataset is available at \href{https://cvl.tuwien.ac.at/research/cvl-databases/h2opm-dataset/}{https://cvl.tuwien.ac.at/research/cvl-databases/h2opm-dataset/}}. For each reference image, 3-11 historical aerial images captured between May 1943 and May 1945 are available, leading to a total of 42 image pairs. An overview of the test data is given in Table~\ref{tab:data}. All images have been scale-normalized to a spatial resolution of 1m prior to processing. In order to evaluate the registration accuracy, manually selected ground truth correspondences between the historical images and the OPM are used and the performance is measured by the Root Mean Squared Error (RMSE):

\begin{equation}
	RMSE = \sqrt{\frac{1}{M}\sum_{i=1}^{m}(x_i-x_i^\prime)^2 + (y_i-y_i^\prime)^2},
\end{equation}

\noindent where $(x_i,y_i)$ and $(x_i^\prime,y_i^\prime)$ are the coordinates of the $M$ ground truth correspondences in the reference OPM and the transformed historical image, respectively. The ground truth correspondences have been selected by GIS experts with the aim of manually georeferencing the historical images. Per historical-OPM image pair, between 4 and 19 correspondences are available, with 11.5 correspondences on average. Ground truth correspondences are visualized for two image pairs in Figure \ref{fig:inlier_ratio_examples}.

\begin{table}[t!]
  \centering 
    \caption{Overview of test data.}
  \label{tab:data}
	\footnotesize
  \begin{tabular}{|p{0.9cm}|p{2.5cm}|p{1.6cm}|p{1.0cm}|p{3.9cm}|}
	\hline
	\textbf{Image set} &\textbf{Location} & \textbf{Area of OPM (m)} & \textbf{\# \newline images} & \textbf{Acquisition interval (days)}\\
	\hline
  1&Vienna District 3& $4990\times 5392$ & 4 & 365/119/36 \\
  2&Linz&$4818\times 4074$ & 5 & 430/233/54/24\\
	3&\"Otztal& $4534\times 4314$& 3& 61/23\\
	4&Schwechat& $5470\times 5234$& 11& 0/0/137/0/76/63/163/0/105/5\\
	5&St. P\"olten&$4050\times 4173$ & 3& 131/114\\
	6&Vienna Lobau& $4373\times 4173$& 3& 130/37\\
	7&Vienna District 21& $5861\times 5123$& 10& 243/72/4/27/59/25/42/35/13\\
	8&Vienna District 23 &$2717\times 2086$ & 3& 61/275 \\
	\hline
\end{tabular}
\end{table}

Since our method is non-deterministic due to PSO and its random initialization, all our reported results are averaged over 50 test runs. However, the standard deviation of the 50 different solutions lies at 0.9m for translation and $0.2^\circ$ for rotation, proofing the high stability of the probabilistic groupwise optimization process.

\subsection{Performance of standard local feature matching}
\label{sec:standard}

In order to motivate the use of our proposed groupwise registration procedure, we investigate the uncertainty of determined local correspondences between  historical images and OPMs. In this experiment, we evaluate the inlier ratio when matching local descriptors on the historical image data, namely SIFT~\cite{Lowe04} and SURF~\cite{Bay08}, which are widely used descriptors in remote sensing image registration \cite{Ma15,Song14,Bouchiha15}. The inlier ratio is defined as the fraction of  matched features in the overall set of putative matches that are actually correct. We follow the standard distance ratio matching method \cite{Lowe04} that accepts a match only if it shows a certain level of unambiguousness in the overall sampled descriptor set, measured by the ratio of distances between the nearest and second nearest descriptor extracted from the reference image. This ratio is thresholded with a defined value $\tau$ ($\tau$=0.7 in our case) to decide for the acceptance of a given match.

\begin{figure}[ht!]
	\centering
\includegraphics[width=0.55\columnwidth]{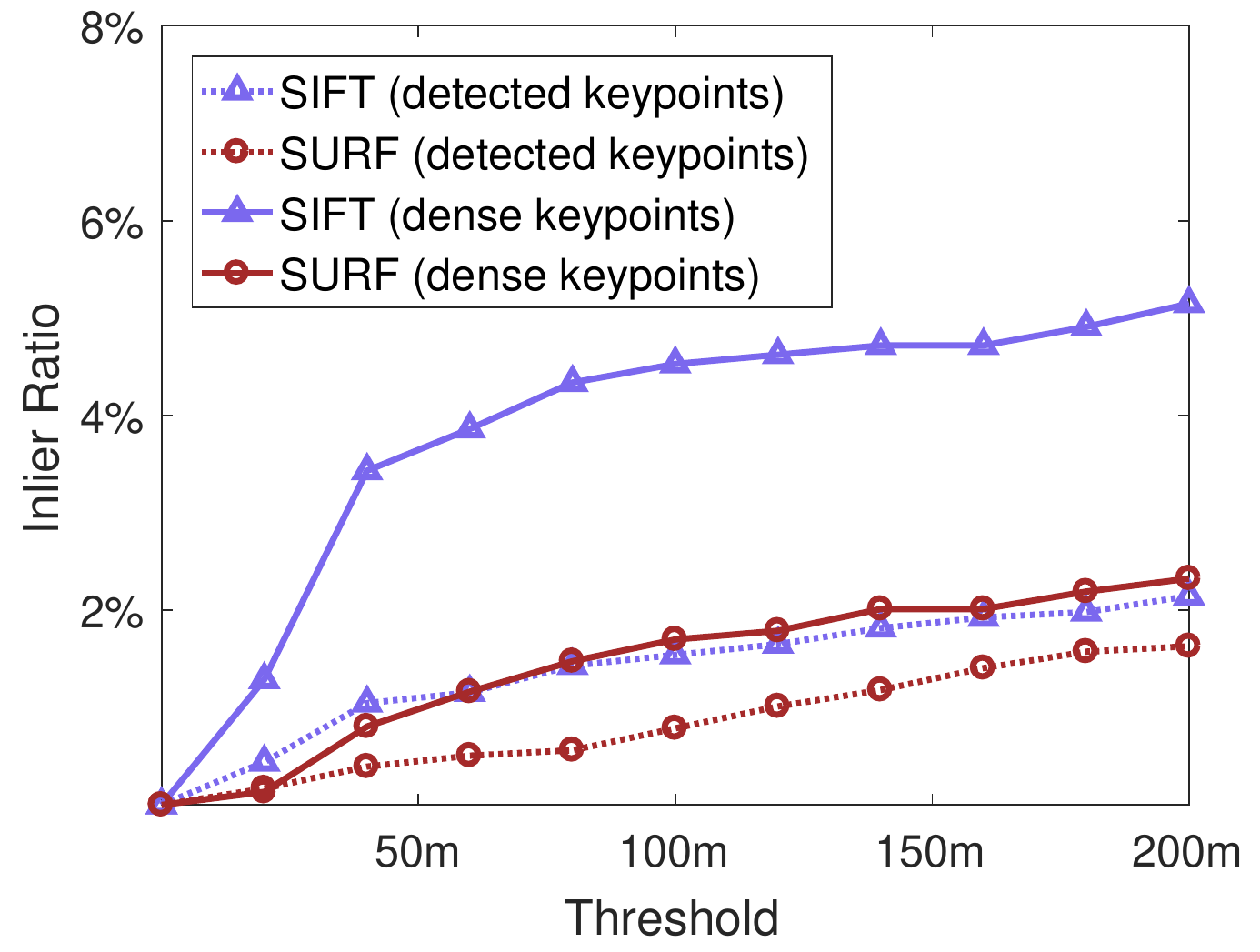}
		\caption{Inlier ratio of detected and densely sampled SURF and SIFT features as a function of inlier acceptance threshold.}
		\label{fig:inlier_ratio}
\end{figure}

In Figure \ref{fig:inlier_ratio}, we show the inlier ratio as a function of the allowed ground distance that determines if a match is considered as correct, evaluated on all 42 pairs of historical-OPM images. Both SIFT and SURF descriptors are tested either with detected or regularly sampled (dense) keypoints \cite{Chatoux16}. This way, the influence of keypoint detection to the matching performance is investigated, but additionally parameters for the step size on the regular grid as well as for the constant scale of the keypoints need to be defined. Based on empirical tests, the results shown in Figure \ref{fig:inlier_ratio} are achieved with a step size of 40m and a local support region of $240\times240$m.

\begin{figure}[t!]
	\centering
	(a)\includegraphics[width=0.75\columnwidth]{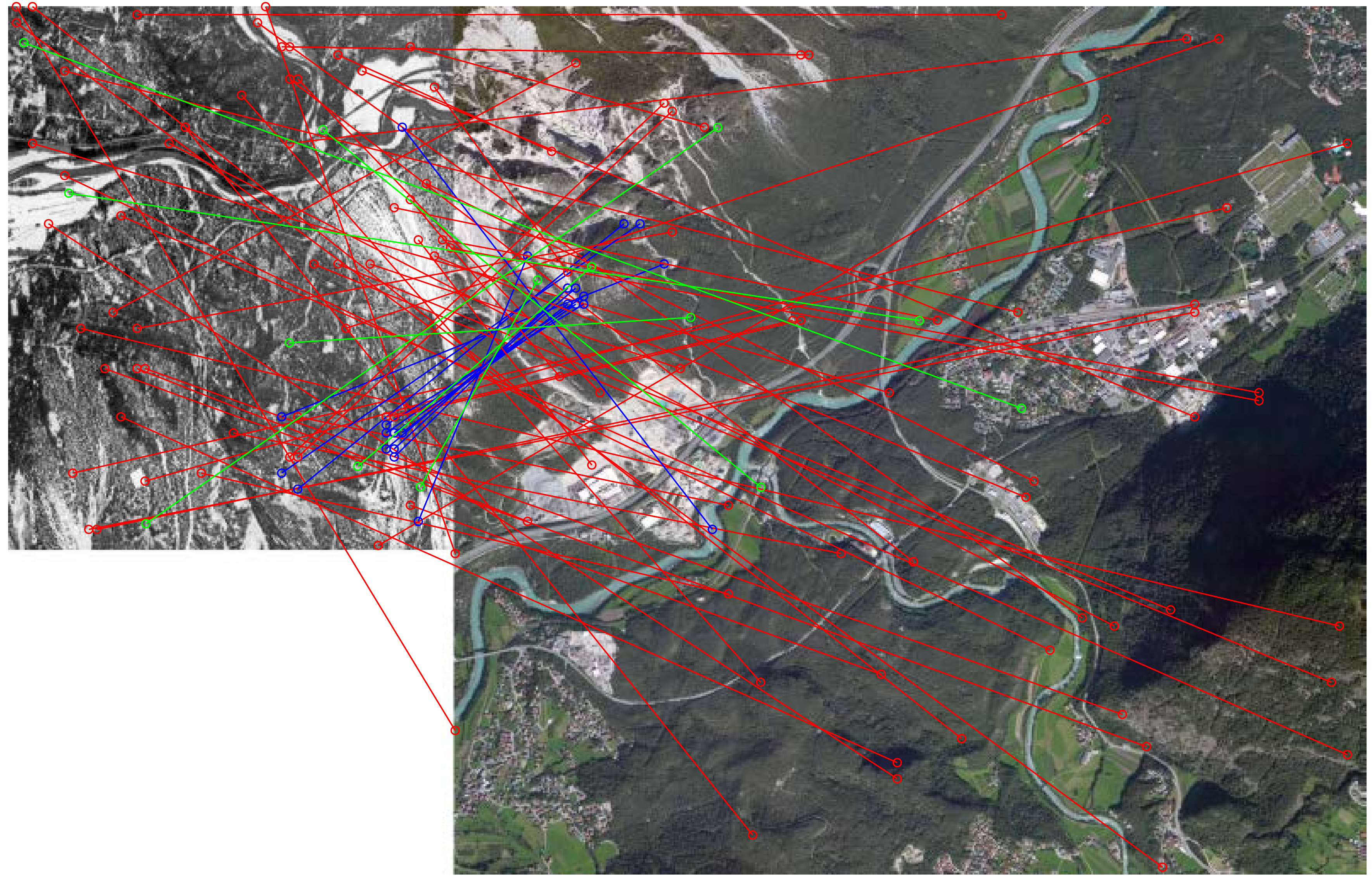}\\
(b)\includegraphics[width=0.75\columnwidth]{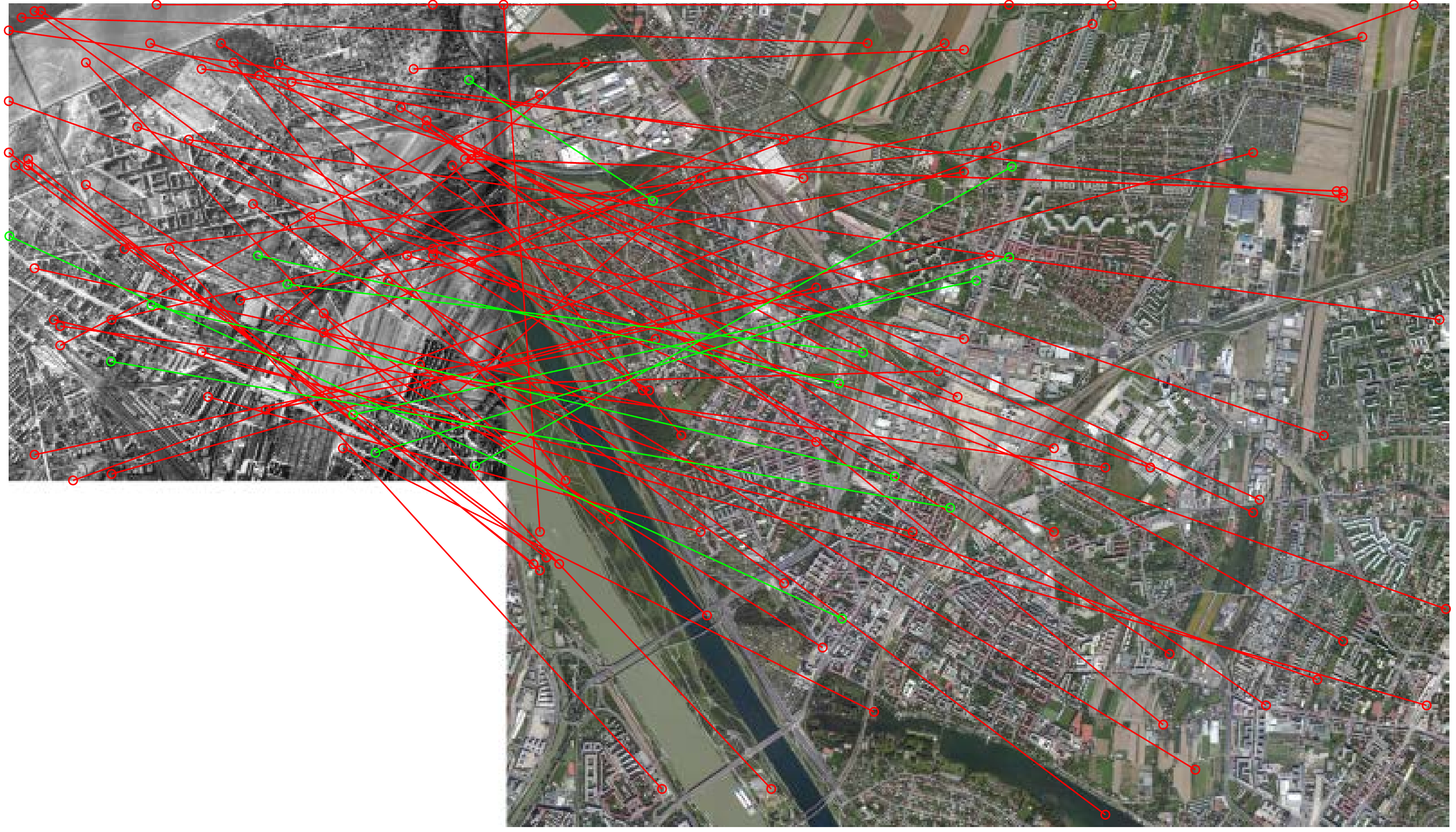}
		\caption{(a) best (18\%) and (b) worst (0\%) inlier ratio achieved on our test with dense SIFT keypoints and a 50m threshold (red: incorrect matches, blue: correct matches, green: manually selected ground truth matches).}
		\label{fig:inlier_ratio_examples}
\end{figure}

It can be concluded from the results that SIFT performs better on this image data than SURF. Additionally, keypoint detection shows an unreliable repeatability and regular sampling is thus the better option. However, most notably, mean inlier ratios are still below 5 $\%$, which is dramatically low compared to other registration scenarios. Even the best performing image pair shows an inlier ratio of only 18\% (see Figure \ref{fig:inlier_ratio_examples}(a)), and for 27 out of 42 images (like the one shown in Figure  \ref{fig:inlier_ratio_examples}(a)) no single correct match can be detected. Typical inlier ratios reported in literature with the same parameter settings are in the order of $75\%$ \cite{Ma18}. The high number of mismatches is attributed to the vast structural differences in image content, and not to a small overlap between images, which is typically a major cause of matching errors \cite{Liu17}. As a conclusion, these results demonstrate that independent, unconstrained pairwise registrations are highly unreliable on this kind of data, and motivate to approach the task with a probabilistic, groupwise registration framework.

\subsection{Groupwise vs. pairwise registration}
\label{sec:comparison1}

In this section, we investigate the performance of our proposed groupwise registration method on the test data. For all tests, we use the best performing variant of dense SIFT feature extraction described in Section \ref{sec:standard} to build the Hough voting spaces between all pairs of images. These Hough voting spaces are then used as pairwise registration estimators in our proposed probabilistic groupwise registration framework to finally estimate the rigid transform of any historical image to the OPM. 

\begin{figure*}[t!]
	\centering
	(a)\includegraphics[height=5.5cm]{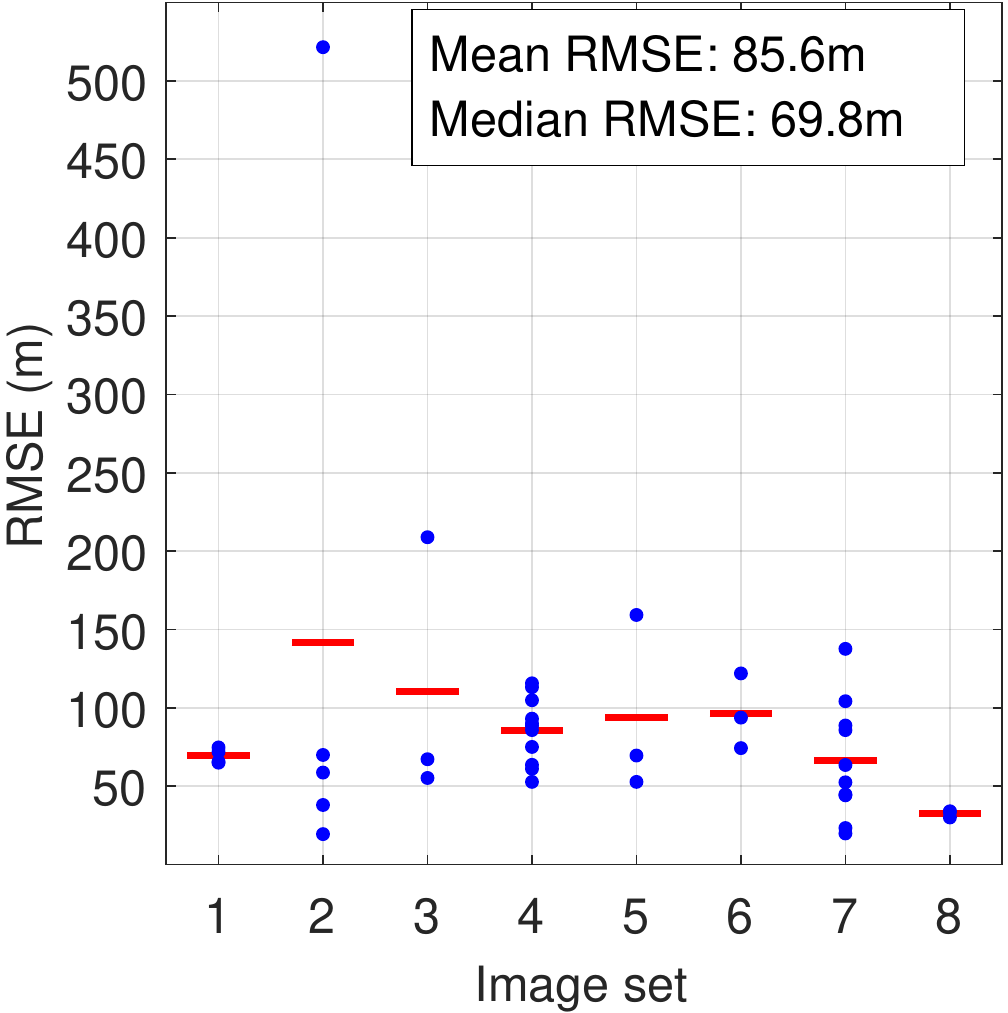}
(b)\includegraphics[height=5.5cm]{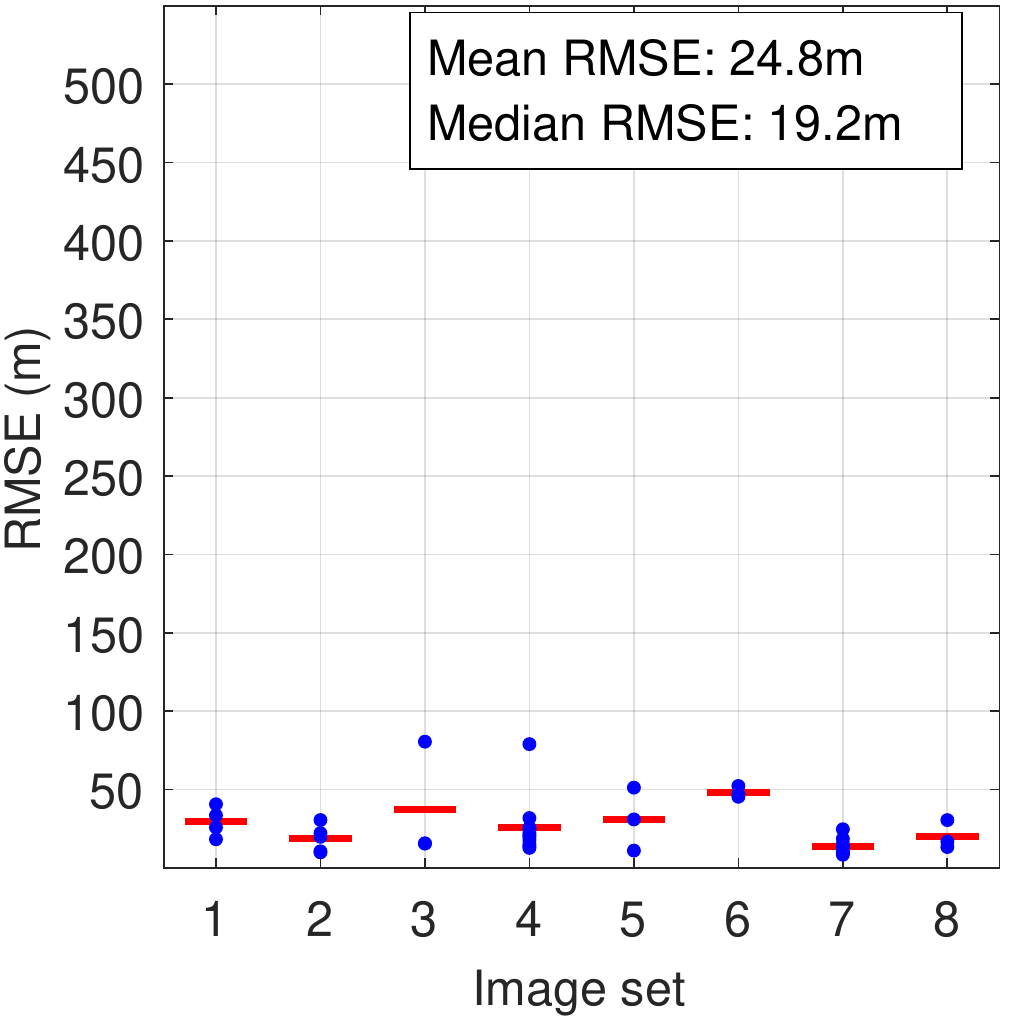}\\
		\caption{RMSE of our proposed groupwise registration method, (a) without and (b) with final guided matching step. A blue point depicts the error of an individual image and red lines depict set mean.}
		\label{fig:rmse_guidedmatching}
\end{figure*}

\begin{figure*}[t!]
	\centering
(a)\includegraphics[height=5.3cm]{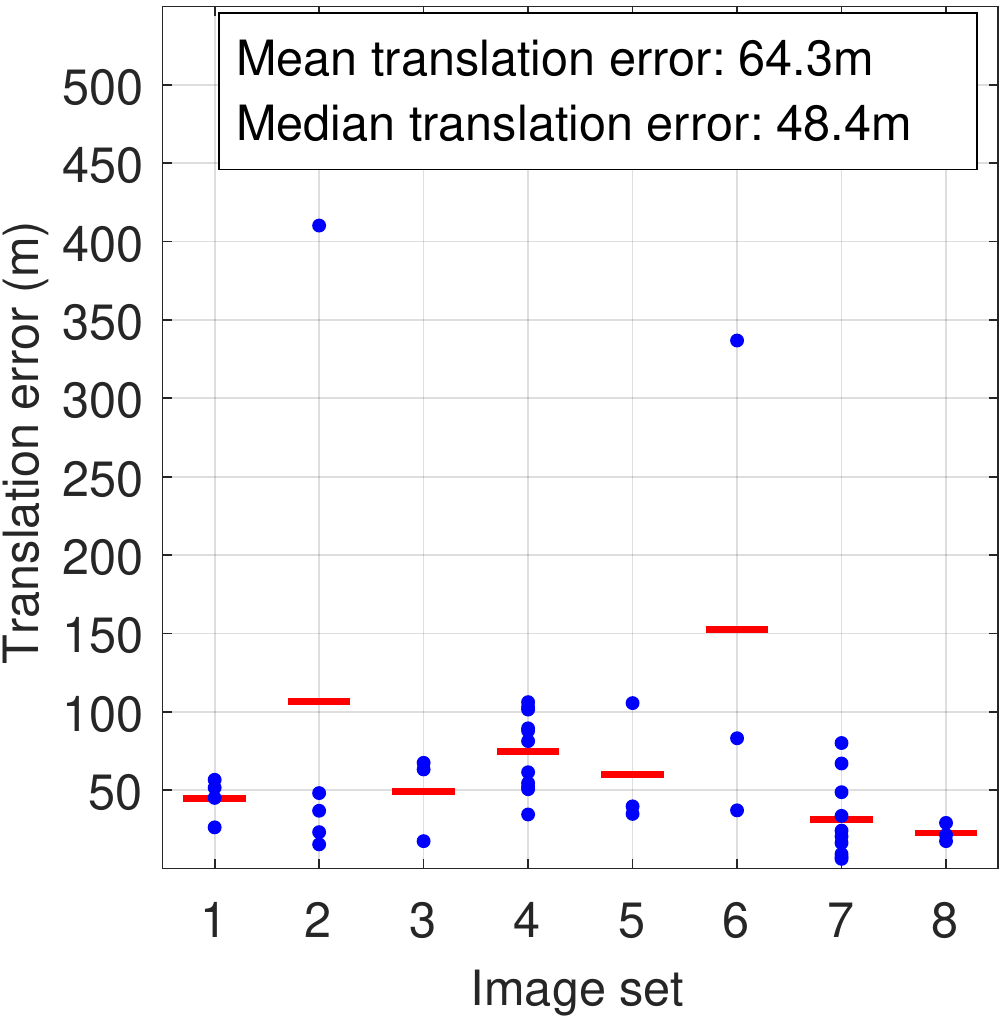}
(b)\includegraphics[height=5.3cm]{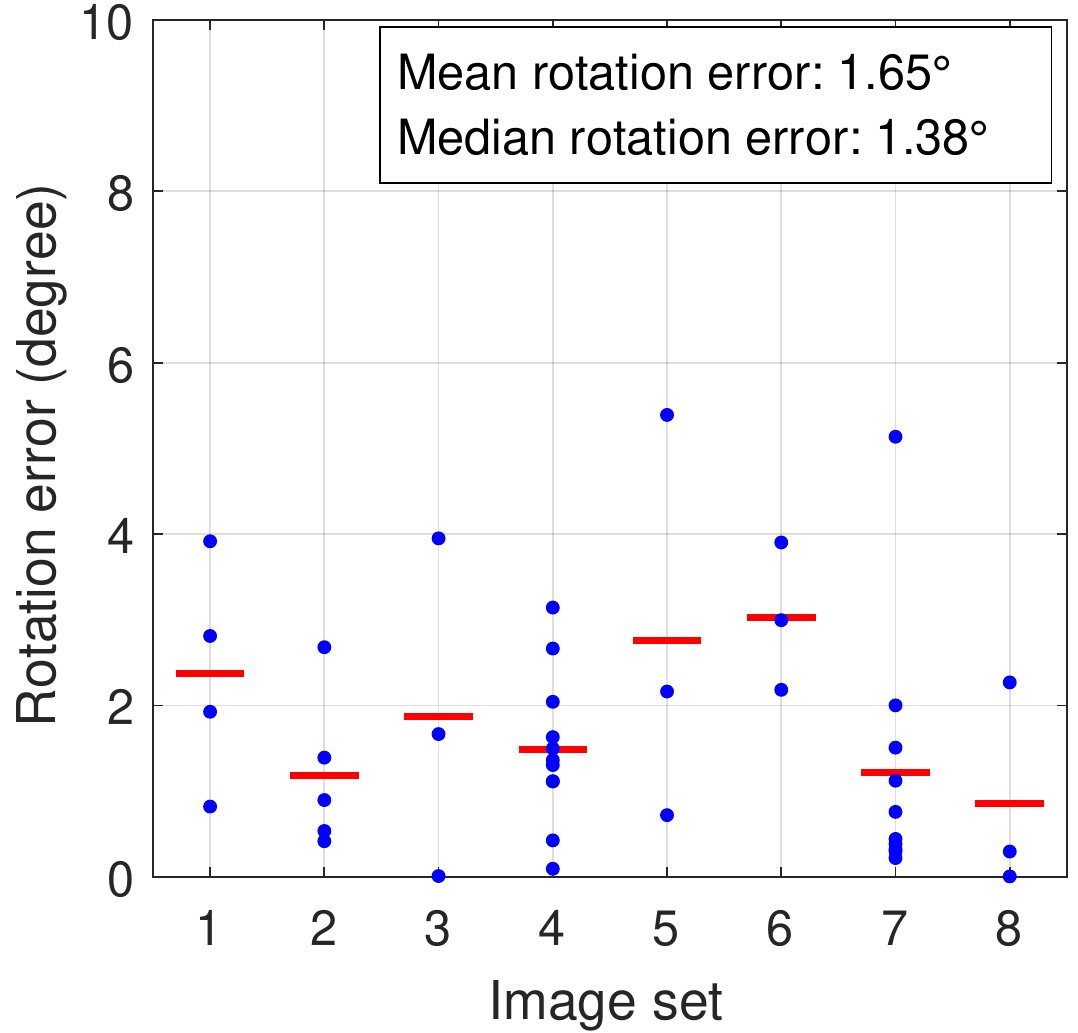}
		\caption{Error components of rigid transformations estimated with our proposed groupwise registration method; (a) error of translation component, (b) error of rotation component.}
		\label{fig:rmse_translation_rotation}
\end{figure*}

Figure \ref{fig:rmse_guidedmatching}(a) shows the RMSE of the registration of all 42 historical images to their respective OPM, when directly taking the estimated rigid transform without a final guided matching step. It can be seen that all eight sets can be registered with a mean RMSE of less than 150m for all sets. When measuring the error of the translation and rotation component individually, as shown in Figure \ref{fig:rmse_translation_rotation}, it becomes evident that errors are mainly attributed to translation estimation, as rotation errors are only $1.64^\circ$ on average. A closer investigation of the outliers with extraordinary large errors (such as the image of set 2 with a RMSE of 521.6m) reveals that high error rates are correlated with remaining scale differences between the images after scale normalization. For instance, for this particular image the estimated scale is $29.9\%$ off the real value due to the inaccurate scale estimation of historical images. On average, the provided scales of the historical images have an error of $4.7\%$ on our test data, with $30.2\%$ being the maximum.

Evidently, for large scale estimation errors the rigid transform assumption is strongly violated and needs to be compensated in the final guided matching step described in Section \ref{sec:guidedmatching}. The results of this step are shown in Figure \ref{fig:rmse_guidedmatching}(b). It can be seen that on all image sets the RMSE is decreased, on average from 85.6m to 24.8m. This shows that guided matching is a robust process and, while especially helpful under large scale differences, provides a generally more accurate registration result.

A major claim of this paper is that the groupwise paradigm is highly beneficial over simple pairwise historical-to-OPM registration. In order to support this claim, we compare our groupwise registration method to its pairwise variant where only the maximum in each historical-to-OPM Hough space is taken, i.e. using only the direct relations term in Eqn. \ref{eq:J} and neglecting inter-historical relations. In addition, we test several pairwise image-to-image registration methods proposed in literature: \emph{SIFT+RANSAC}, \emph{Locally Linear Transforming (LLT)}~\cite{Ma15} and \emph{Progressive Graph Matching (PGM)}~\cite{Cho12}. SIFT+RANSAC corresponds to the dense SIFT feature matching and RANSAC outlier removal. LLT is a maximum likelihood method for the concurrent transformation estimation and outlier removal for a set a tentative matches by means of a locally linear constraint. PGM is an iterative graph matching algorithm that considers both feature similarity and geometric consistencies to find optimal feature correspondences. Both LLT and PGM rely on an initial set of tentative matches, which are provided by densely extracted SIFT features in our case. Compared to the best performing SIFT matching method described in Section \ref{sec:standard}, we use a slightly different parameter set of $360\times360$m local support region and $\tau=1.3^{-1}$ as it performed better with RANSAC, LLT and PGM.

\begin{figure*}[t!]
	\centering
\includegraphics[width=0.7\textwidth]{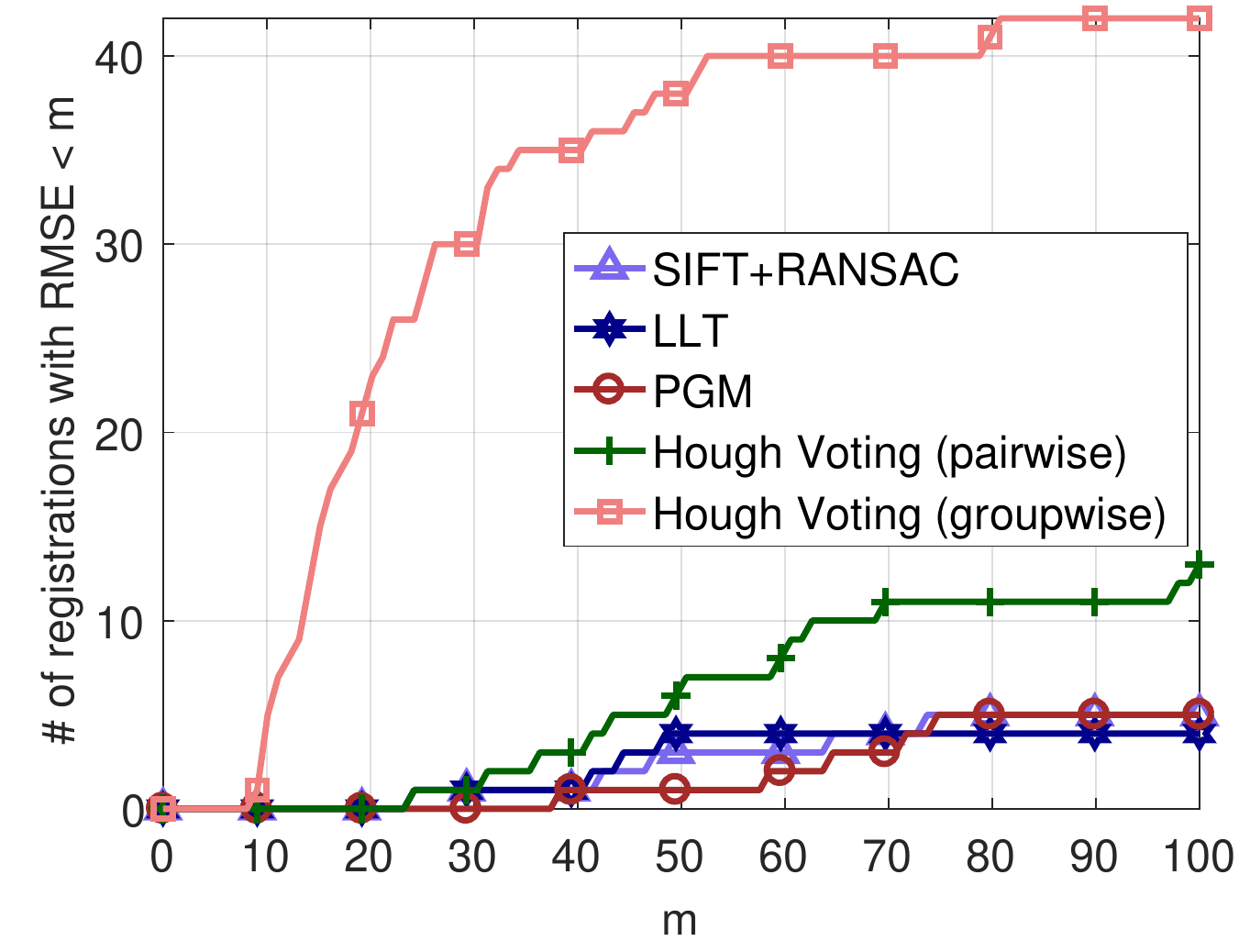}\\
		\caption{Number of correct registrations as a function of RMSE threshold for our groupwise registration method and competing pairwise registration methods.}
		\label{fig:comparison1}
\end{figure*}

The results of all tested methods are provided in Figure \ref{fig:comparison1}, by means of the number of correct registrations as a function of RMSE threshold. It can be seen that the performance of conventional pairwise registration methods is bounded. This boundary is defined by the inevitable vast differences in image content resulting in very unreliable feature similarity estimations and can thus also not simply be crossed by applying more sophisticated matching schemes. Therefore, a meaningful registration for all test images can only be achieved by considering their geometric interrelations, as done by our proposed probabilistic groupwise registration framework.

\begin{figure}[ht!]
	\centering
(a)\begin{minipage}[b]{.29\textwidth}\includegraphics[width=\textwidth]{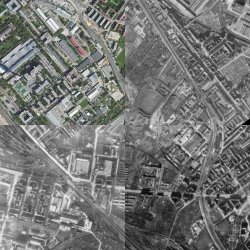}\vspace{0.1cm}
						\includegraphics[width=\textwidth]{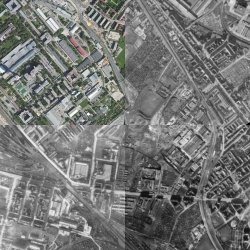}\end{minipage}
(b)\begin{minipage}[b]{.29\textwidth}\includegraphics[width=\textwidth]{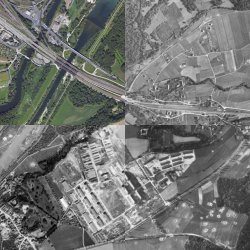}\vspace{0.1cm}
						\includegraphics[width=\textwidth]{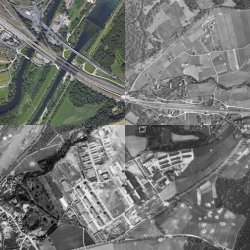}\end{minipage}
(c)\begin{minipage}[b]{.29\textwidth}\includegraphics[width=\textwidth]{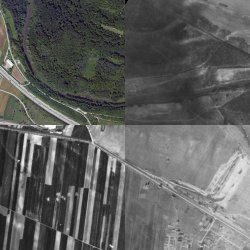}\vspace{0.1cm}
						\includegraphics[width=\textwidth]{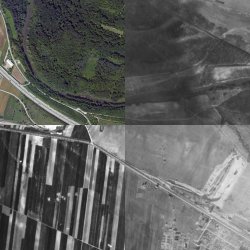}\end{minipage}	
(d)\begin{minipage}[b]{.29\textwidth}\includegraphics[width=\textwidth]{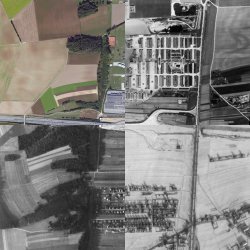}\vspace{0.1cm}
						\includegraphics[width=\textwidth]{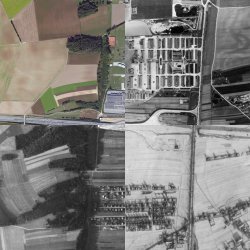}\end{minipage}
(e)\begin{minipage}[b]{.29\textwidth}\includegraphics[width=\textwidth]{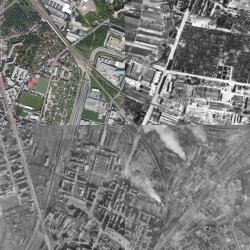}\vspace{0.1cm}
						\includegraphics[width=\textwidth]{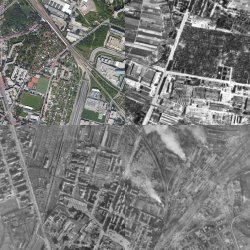}\end{minipage}
(f)\begin{minipage}[b]{.29\textwidth}\includegraphics[width=\textwidth]{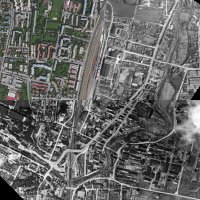}\vspace{0.1cm}
						\includegraphics[width=\textwidth]{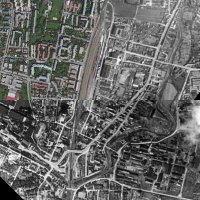}\end{minipage}	

			\caption{Historical-to-OPM registration results. Each mosaic consists of a reference region (OPM, top left corner of the image) and three aligned historical images. Top mosaics correspond to ground truth, bottom mosaics have been created with our proposed groupwise registration method. Mosaics (a)-(f) are from image sets 1, 2, 4, 5, 7 and 8, respectively. }
		\label{fig:results_images}
\end{figure}

In Figure \ref{fig:results_images}, exemplary registration results achieved with our method are shown. Visually, the results widely resemble the ground truth registrations. Inaccuracies can be spotted for especially hard cases like the snow-covered landscape at the bottom right corner of Figure \ref{fig:results_images}(d). While it is largely difficult to assess the registration accuracy from the mosaics due to the heavily changing image content between historical and OPM images, street courses seem to represent the most constant structure in the images. Still, their closer inspection suggests that the ground truth manually achieved by analysts is also not perfectly accurate, as partly in the automatically achieved mosaics street courses fit better and thus the automatic result seems to be more reasonable (cf. Figure~\ref{fig:results_images}(f)).

\subsection{Sequential global optimization vs. direct optimization}
\label{sec:optimization_results}

The global optimization procedure presented in Section \ref{sec:optimization} aims at optimizing Eqn. \ref{eq:J} in a sequential manner in order to avoid local maxima trapping. In this section, we demonstrate the superior convergence behavior of this optimization strategy compared to directly optimizing for all function parameters simultaneously. For this purpose, we apply PSO to the function with different numbers of particles (10, 100 and 1000). All experiments are conducted 10 times with randomly chosen initial swarm populations. The initial swarm populations are sampled from discrete probability distributions represented by the direct relations $H^{k,*}(\mathbf{T}_k)$ in order to favor initializations with a high probability in the historical-OPM registration estimators, which are likely closer to the global maximum. 

\begin{figure*}[b!]
	\centering
	(a)\includegraphics[height=4.3cm]{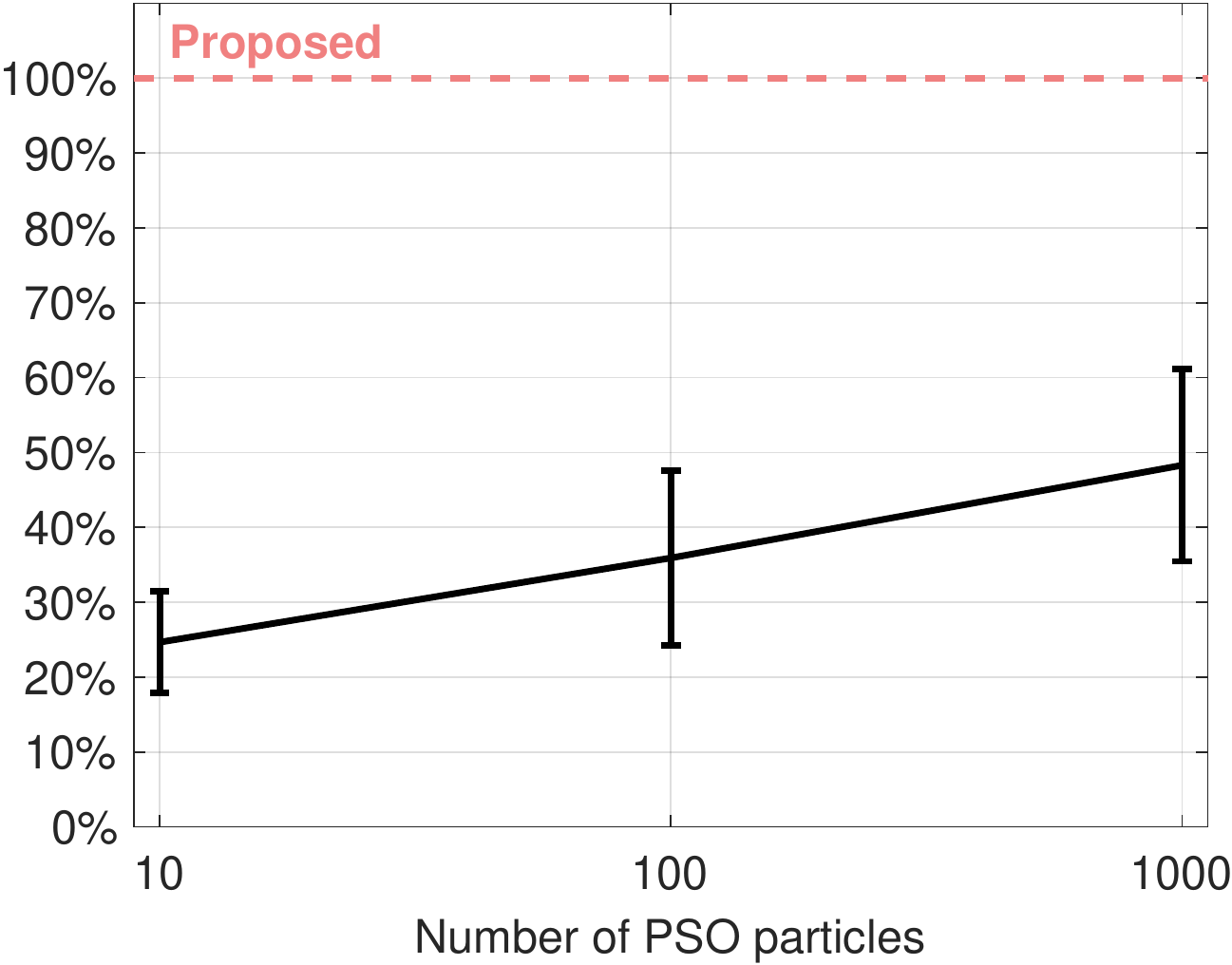}
(b)\includegraphics[height=4.3cm]{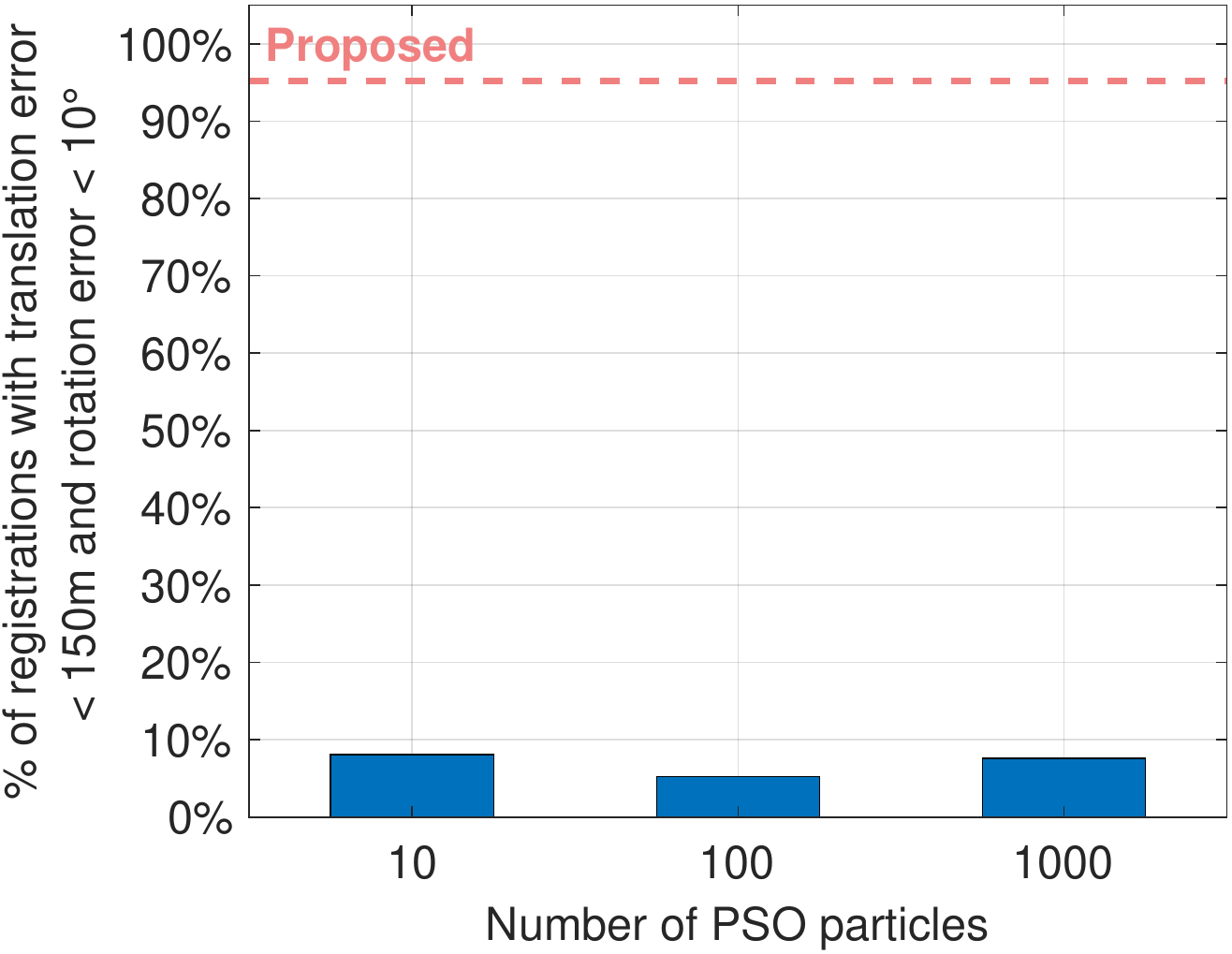}\\
		\caption{(a) Mean and standard deviations of the final fitness function values achieved according to the number pf PSO particles used, relative to the function values achieved by our proposed sequential optimization, (b) percentage of historical-OPM registrations with a translation error of less than 150m and a rotation error of less than $10^\circ$. The red dotted line indicates the performance of our proposed sequential optimization procedure (cf. Figure \ref{fig:rmse_translation_rotation}). }
		\label{fig:optimization}
\end{figure*}

In terms of the final fitness function value achieved, adding more particles helps to reach a solution that is on average closer to the one achieved by our sequential optimization procedure. This can be seen in Figure \ref{fig:optimization}(a) where the means and standard deviations of the PSOs with different number of particles are plotted. However, in terms of finding the correct solution, these particle numbers have no conceivable influence. This is demonstrated in Figure \ref{fig:optimization}(b) where the percentages of solutions with a translation error of less than 150m and a rotation error of less than $10^\circ$ are shown. Although using even more particles would certainly contribute to higher success rate, it would come at the cost of having a longer processing time. As can be seen in Table  \ref{tab:optimization_time}, the average processing time with 1000 particles is already 2.6 times longer than the processing time of our proposed sequential optimization. 

\begin{table}[t]
  \centering 
    \caption{Processing times in seconds (mean and standard deviation) of our proposed sequential optimization scheme and direct PSO with 10, 100 and 1000 particles.}
  \label{tab:optimization_time}
	\footnotesize
	\begin{tabular}{|p{1.1cm}|p{0.5cm}p{1cm}|p{0.5cm}p{1cm}|p{0.5cm}p{1cm}|p{0.5cm}p{1.2cm}|}
	\hline
	\textbf{Image set} & \multicolumn{2}{l|}{\textbf{Proposed}} & \multicolumn{2}{l|}{\textbf{PSO (10)}} & \multicolumn{2}{l|}{\textbf{PSO (100)}} & \multicolumn{2}{l|}{\textbf{PSO (1000)}}\\
	\hline
1 & 98.8 &$\pm$ 7.4 & 5.4 &$\pm$ 1.3 & 34.6 &$\pm$ 18.5 & 366.7 &$\pm$ 182.7 \\         
2 & 137.6 &$\pm$ 4.3 & 6.9 &$\pm$ 1.8 & 53.2 &$\pm$ 18.6 & 517.5 &$\pm$ 400.3 \\ 
3 & 43.3 &$\pm$ 6.1 & 3.3 &$\pm$ 1.3 & 15.8 &$\pm$ 8.0 & 246.9 &$\pm$ 89.3 \\           
4 & 1220.1 &$\pm$ 80.5 & 29.5 &$\pm$ 11.8 & 249.9 &$\pm$ 100.4 & 2488.1 &$\pm$ 1114.1 \\
5 & 50.6 &$\pm$ 6.9 & 2.5 &$\pm$ 0.8 & 16.9 &$\pm$ 10.7 & 199.9 &$\pm$ 139.5 \\         
6 & 47.9 &$\pm$ 4.5 & 2.9 &$\pm$ 0.8 & 16.4 &$\pm$ 8.9 & 274.7 &$\pm$ 137.7 \\          
7 & 959.8 &$\pm$ 70.5 & 29.6 &$\pm$ 16.7 & 185.7 &$\pm$ 240.9 & 2605.5 &$\pm$ 904.3 \\  
8 & 49.7 &$\pm$ 4.6 & 2.3 &$\pm$ 0.5 & 17.7 &$\pm$ 8.9 & 205.1 &$\pm$ 121.7 \\
\hline          
Average & 326.0 &$\pm$ 23.1 & 10.3 &$\pm$ 4.4 & 73.8 &$\pm$ 51.9 & 863.0 &$\pm$ 386.2 \\          
\hline
\end{tabular}
\end{table}

\subsection{Comparison to state-of-the-art}
\label{sec:comparison2}

As stated above, related work in groupwise image registration is scarce and most of methods are not applicable to our problem since they are grounded in global image registration and hence rely on a too restrictive transformation model between images. In order to compare our method to existing state-of-the-methods in multi-image feature-based image registration, we implemented a method that follows the idea of topology estimation for stitching together unordered image sets, as proposed in several works in the past \cite{Xia17,Elibol13,Choe06}. Common to these methods is that the image topology is represented as an undirected graph, where the nodes correspond to the images and the edge weights correspond to an estimated quality metric of the pairwise registration. Given such a graph representation, the goal is to find an optimal reference image and optimal linkage paths to the reference image with minimal registration error, e.g. by solving the minimum spanning tree problem \cite{Graham85} or Floyd Warshall all-pairs shortest path algorithm \cite{Floyd62}. 

Specifically, for comparison we adapted the method of \cite{Xia17} to our data and performed the following steps to obtain a registration of an image set. First, all image pairs are matched by means of densely extracted SIFT features followed by RANSAC spatial verification. The number of matched features is used as a confidence metric for the estimated transformation between two images. Then an undirected graph is generated whose edges are weighted with inverse confidence values. Finally, Floyd Warshall all-pairs shortest path algorithm is applied to the graph to determine the linkage paths from all historical images to the reference OPM.

\begin{figure*}[b!]
	\centering
(a)\includegraphics[height=5.3cm]{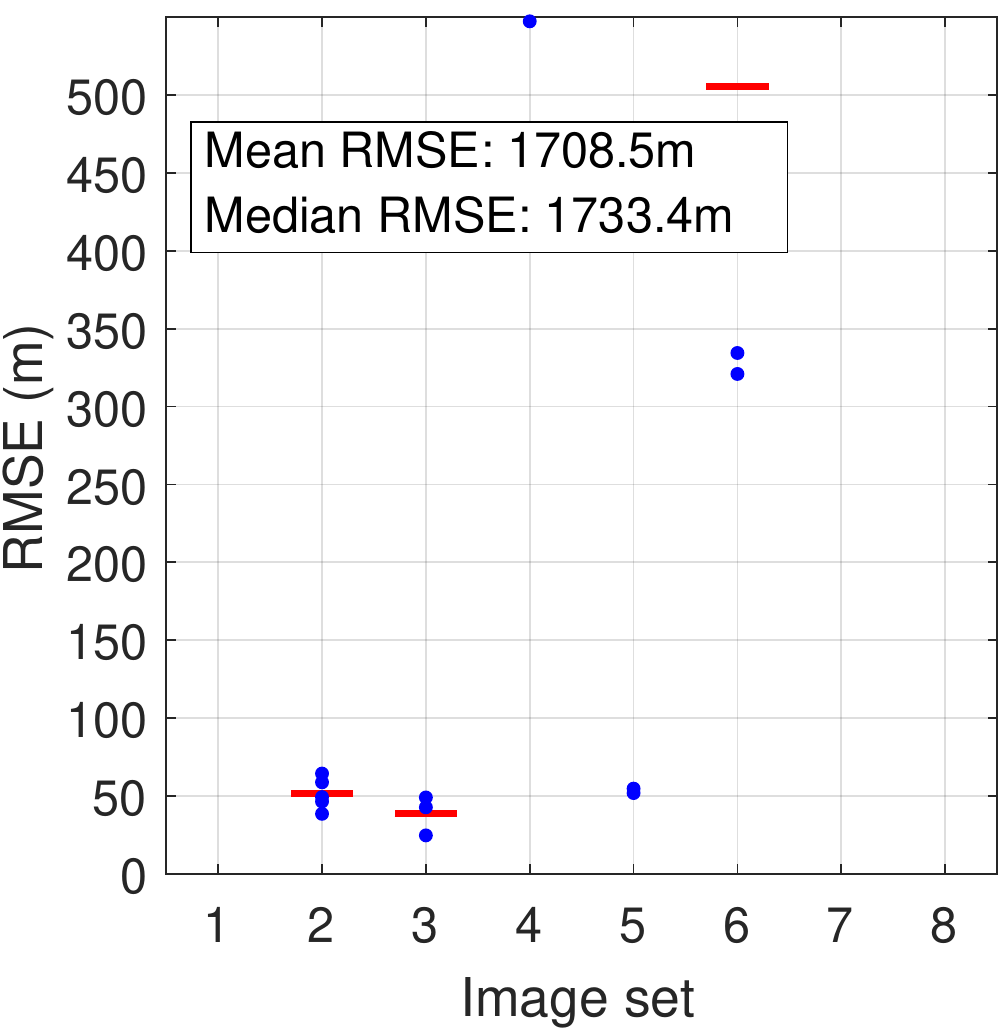}
(b)\includegraphics[width=0.48\columnwidth]{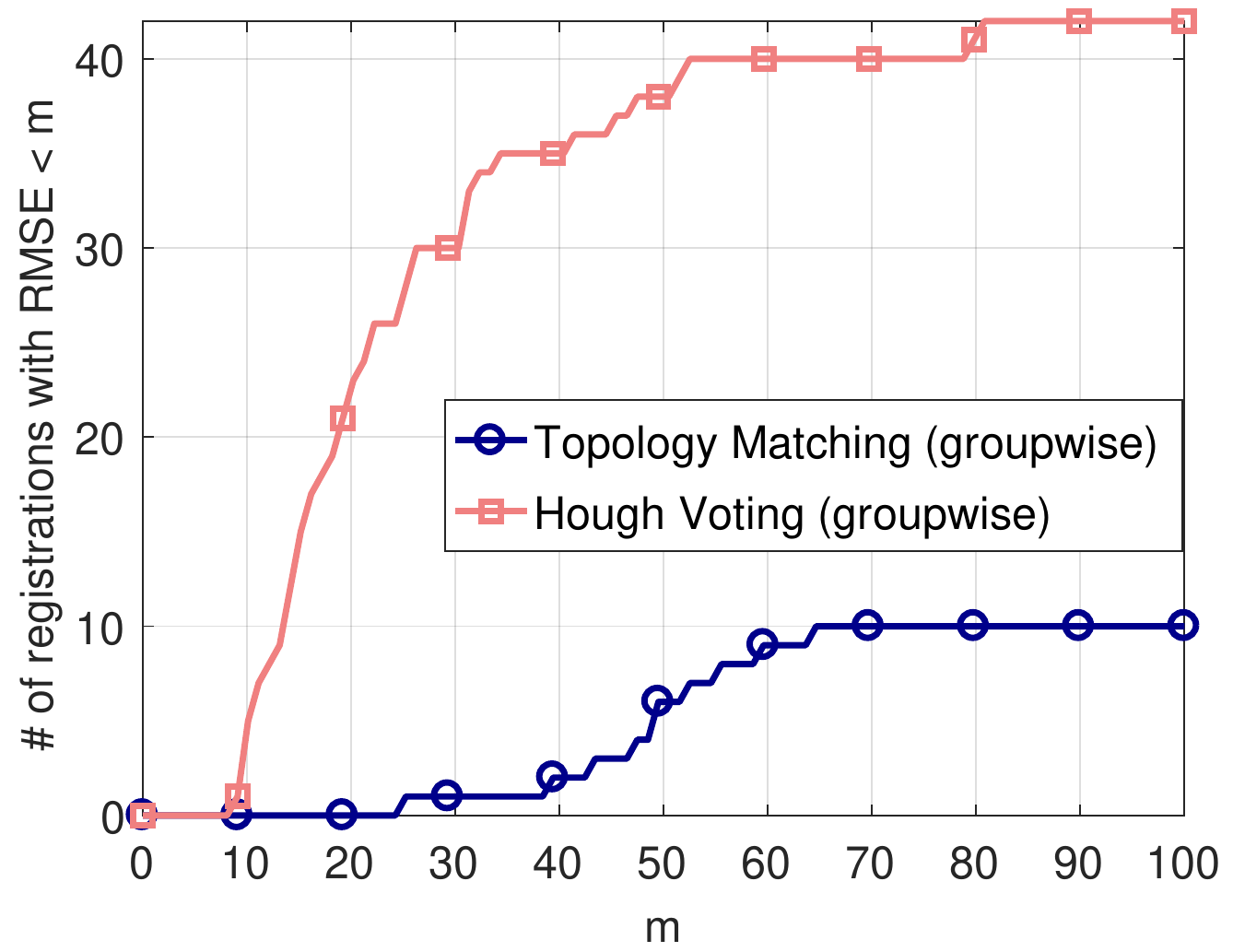}
		\caption{(a) RMSE of groupwise registration via topology matching \cite{Xia17} for all 8 image sets; (b) Comparison of correctly registered images with topology matching \cite{Xia17} and our proposed method.}
		\label{fig:comparison2}
\end{figure*}

From the results shown in Figure \ref{fig:comparison2}, it can be observed that topology matching is only able to obtain a reasonable registration for the image sets 2 and 3, and a similar accuracy is only achieved on image set 3 (38.9m mean RMSE vs. 37.2m mean RMSE of the proposed method). The main problem of topology matching is that transformations between two images are fixed at an early stage, only by considering their pairwise relation. Its primary objective is to minimize the error propagation by estimating and analyzing the topology of the unordered image set, but with the premise that pairwise registrations are reliable enough  to find stable linkage paths from all images to the reference. This is especially problematic for the historical-to-OPM relations that cannot be estimated with the robustness needed.

\section{Conclusions}
\label{sec:conclusions}

In this paper, we introduced a novel probabilistic framework for the groupwise registration of aerial historical images to present-day OPMs. Unlike previous groupwise registration methods that originate from the global image registration world, it is a local feature-based method and thus able to handle arbitrary translational and rotational differences between images, as well as typical scale changes of up to 30\% present in our test data. Other feature-based methods register groups of images in a pairwise manner, and these initial pairwise registrations serve as input for an overall registration refinement by means of techniques such as bundle adjustment or topology analysis. However, this early rough fixation of pairwise registrations increases the risk of overall registration failure in the case of highly unreliable pairwise registrations, and we solve this problem by using a Hough voting space as pairwise registration estimator that allows to determine the max-likelihood joint registration. The Hough voting space can be seen as the fundament of our approach, as it delivers what pixel-based image similarity metrics do for groupwise global registration: providing a probability estimation for each transformation between the images. 

We compiled a very challenging dataset of 42 historical images, representing the manifold characteristics that make historical-to-OPM registration hard to achieve.  Consequently, state-of-the-art pairwise and groupwise registration methods failed to find a reasonable registration for at least 70\% of the images. In contrast, the proposed method succeeded for all 42 historical test images with a maximum and mean error of 80.5m and 24.8m, respectively. For future work, we aim to further improve the fine registration of images, by investigating non-rigid flexible registrations of pre-aligned images \cite{Yang11} as well as by leveraging 3D bundle adjustment techniques that minimize the overall reprojection error~\cite{Konolige10}.

\section*{Acknowledgments}
This work was supported by the Austrian Research Promotion Agency (FFG) under project grant 850695. The authors wish to thank\textit{ Luftbilddatenbank Dr. Carls GmbH}. \\
Acquisition of historical aerial imagery: \textit{Luftbilddatenbank Dr. Carls GmbH}; Sources of historical aerial imagery: \textit{National Archives and Records Administration} (Washington, D.C.) and \textit{Historic Environment Scotland} (Edinburgh).

\section*{References}

\bibliography{Literature}

\end{document}